\documentclass[twoside,leqno,twocolumn]{article}

\usepackage[letterpaper]{geometry}

\usepackage{ltexpprt}
\usepackage{hyperref}
\usepackage{graphicx}%
\usepackage{multirow}%
\usepackage{amsmath,amssymb,amsfonts}%
\usepackage{mathrsfs}%
\usepackage{xcolor}%
\usepackage{textcomp}%
\usepackage{manyfoot}%
\usepackage{booktabs}%
\usepackage{algorithm}%
\usepackage{algorithmicx}%
\usepackage{algpseudocode}%
\usepackage{listings}%
\usepackage{float}
\usepackage{threeparttable}
\usepackage{subfig}
\usepackage{footnote}

\begin{document}

\newcommand\relatedversion{}

\title{\Large Fine-grained Spatio-temporal Event Prediction  with Self-adaptive Anchor Graph}

\author{Wang-Tao Zhou\footnote{School of Computer Science and Engineering, University of Electronic Science and Technology of China. \{wtzhou@std.uestc.edu.cn, zkang@uestc.edu.cn, 202321081425@std.uestc.edu.cn, l.zhang@uestc.edu.cn,\\lingtian@uestc.edu.cn\}}
\and Zhao Kang\footnotemark[1]~\footnote{Shenzhen Institute for Advanced Study, University of Electronic Science and Technology of China.}
\and Sicong Liu\footnotemark[1]
\and Lizong Zhang\footnotemark[1]~\footnotemark[2]
\and Ling Tian\footnotemark[1]~\footnotemark[2]~\footnote{Kashi Institute of Electronics and Information Industry.}
}

\date{}

\maketitle







\begin{abstract} \small\baselineskip=9pt Event prediction tasks often handle spatio-temporal data distributed in a large spatial area. Different regions in the area exhibit different characteristics while having latent correlations. This spatial heterogeneity and correlations greatly affect the spatio-temporal distributions of event occurrences, which has not been addressed by state-of-the-art models. Learning spatial dependencies of events in a continuous space is challenging due to its fine granularity and a lack of prior knowledge. In this work, we propose a novel Graph Spatio-Temporal Point Process (GSTPP) model for fine-grained event prediction. It adopts an encoder-decoder architecture that jointly models the state dynamics of spatially localized regions using neural Ordinary Differential Equations (ODEs). The state evolution is built on the foundation of a novel Self-Adaptive Anchor Graph (SAAG) that captures spatial dependencies. By adaptively localizing the anchor nodes in the space and jointly constructing the correlation edges between them, the SAAG enhances the model’s ability of learning complex spatial event patterns. The proposed GSTPP model greatly improves the accuracy of fine-grained event prediction. Extensive experimental results show that our method greatly improves the prediction accuracy over existing spatio-temporal event prediction approaches.\end{abstract}

\section{Introduction}
The prediction of spatial-temporal events has become an important task in many applications, such as earthquake prediction \cite{earthquake}, crime prevention \cite{crime}, spacecraft anomaly detection \cite{spacecraft}, and epidemic control \cite{epidemic}. By accurately anticipating the time and location of future events, we can avoid potential risks or dangers and maximize benefits. With the rapid development of deep learning, event prediction techniques have been extensively studied \cite{zhao2015,gao2018,gao2019,staple,wang2021,yu2023}. These works discretize space and time in order to simplify the problem into normal classification or regression tasks, which can be easily solved with deep neural networks. However, they fail to make accurate fine-grained predictions of the arrival times and locations of future events, which limits their application in many practical scenarios.

Spatio-Temporal Point Process (STPP) models have become a research hotspot for fine-grained event prediction problems lately. 
Recent STPP works propose to jointly model the continuous spatio-temporal event distributions with deep neural networks. NJSDE \cite{njsde} adotps a neural Ordinary Differential Equation (ODE) to model the spatio-temporal state transition. NSTPP \cite{nstpp} inherits the ODE-based architecture but incorporates conditional normalising flows to generate a more flexible spatial distribution. DeepSTPP \cite{deepstpp} uses Variational Auto Encoders (VAEs) to model the joint intensity function. DSTPP \cite{dstpp} applies the conditional denoising diffusion technique to the generation of spatio-temporal events. These methods take the spatial locations of the events as ordinary time series and ignore the heterogeneity and correlations between different spatial regions.


Spatial heterogeneity means that different spatial regions possess different patterns of event occurrence at one time. Thus, it is more reasonable to model the state transitions of different local regions separately than globally. Spatial correlations attend to the message passing between different regions. The state evolutions of different spatial regions are not independent. The state change of one region may be largely affected by other regions. Hence a correlation graph is needed to encode such interdependencies. Existing STPP methods mostly use a global state vector to model the event dynamics, without region-specific considerations, thus failing to capture the spatial pattern of event occurrences accurately.

However, localized state modeling is challenging because of two issues. First, the possible locations in a continuous space are infinite and there are no explicit borders that split the whole area into regions. Second, no prior knowledge is available to construct the characteristics and inter-correlations of the regions.

\begin{figure*}[!ht]
    \centering
    \includegraphics[width=\linewidth]{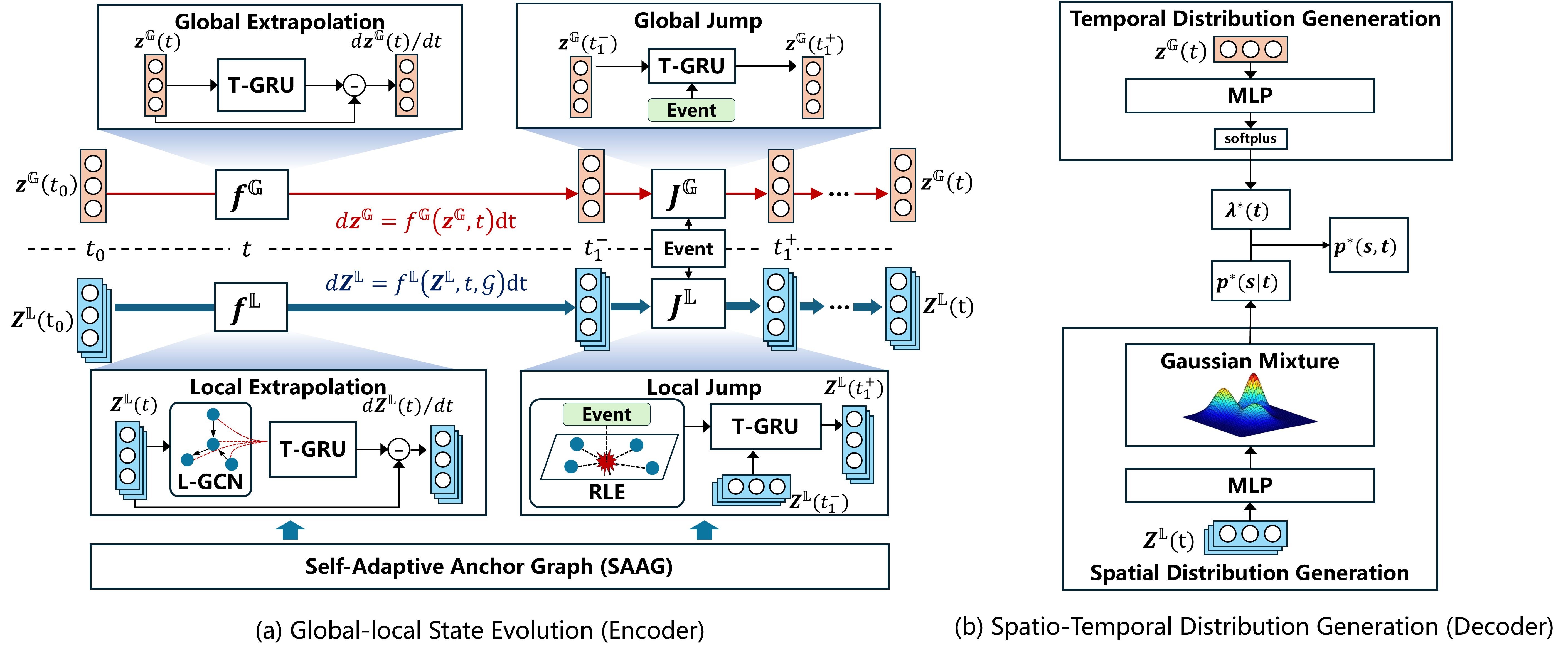}
    \vspace{-2em}
    \caption{The overall framework of the proposed GSTPP model. We adopt an encoder-decoder structure. The encoder simulates the global-local state trajectories, while the decoder generates the spatio-temporal event distribution.}
    \vspace{-1em}
    \label{fig:framework}
    \vspace{-1em}
\end{figure*}

In this work, we propose a novel Graph Spatio-Temporal Point Process (GSTPP) framework to overcome these issues. The overview of the model is shown in Fig. \ref{fig:framework}. Apart from a global state $ \boldsymbol{z}^\mathbb{G} $ that models the overall event occurrence rate, we construct a set of local states $ \boldsymbol{Z}^\mathbb{L} $ to capture the dynamics of different spatial regions. The evolution of the states is simulated with neural ODEs with gated jumps. The model follows an encoder-decoder architecture. The encoders simulate the global-local state trajectories along the temporal axis, while the decoders generate the predicted spatio-temporal distributions of the target events. Inspired by \cite{lmvsc,cdc}, we propose a novel Self-Adaptive Anchor Graph (SAAG) to overcome the challenge of localized state modeling and address the two issues mentioned above.
 Instead of splitting the area with hard borders, SAAG adaptively localizes its anchor nodes in the space that capture the dynamics of their nearby areas. The correlation edges between the anchors are also adaptively learned during training, eliminating the need for prior knowledge. Several submodules like the Location-aware Graph Convolutional Network (L-GCN) and Relative Location Encoder (RLE) are also devised to leverage the spatial pattern learned by SAAG to boost event prediction performance. The contribution of this work can be concluded as follows.

\vspace{0.5em}
\begin{itemize}
    \item We propose GSTPP framework, which jointly models the evolution of global state and region-specific local states. The introduction of localization enables the model to comprehensively learn the complex spatial patterns of different regions, facilitating accurate spatial predictions. To the best of our knowledge, our work is among the first attempts to incorporate spatially localized dynamics for spatio-temporal event prediction.
    
    \item We devise a Self-Adaptive Anchor Graph (SAAG) to address the issue of spatial heterogeneity and correlations. The graph is capable of self-adaptive localization and correlation learning. Using several novel submodules, such as L-GCN and RLE, we utilize the spatial pattern learned from SAAG to boost fine-grained event prediction performance.

    \item Extensive experiments are performed to validate the superiority of the proposed GSTPP model over state-of-the-art STPP models in spatio-temporal event prediction tasks.
    
\end{itemize}


\vspace{0.5em}
\section{Related Works}
\label{sec:related}
\subsection{Spatio-temporal Event Prediction}
Extensive research has been conducted on the prediction of spatio-temporal events. Jointly modeling the continuous spatio-temporal event pattern is challenging; thus, it is a common practice to discretize the time and space in event prediction tasks. Works like LASSO \cite{zhao2015}, MITOR \cite{gao2018}, and SIMDA \cite{gao2019} treat spatio-temporal event prediction as a multi-task learning problem, where the time is cut into windows and each location is taken as a separate event prediction task. Other works like STAPLE \cite{staple}, STCGNN \cite{wang2021}, and STEP \cite{yu2023} propose to learn the spatial correlations by organizing the different locations into a graph, where each node represents a country or city. Although these works jointly consider the times and locations of events, they can only model the locations as discrete labels instead of Euclidean coordinates and hence fail to fit in scenarios where fine-grained spatial prediction is required. Traffic-related event prediction models either divide the Euclidean space into boxes \cite{yuan2018,jin2022} or construct a fixed road network \cite{fang2021,li2023}. However, the accuracy of the spatio-temporal prediction still depends on the granularity they choose and cannot adapt to sparse distributions.

\subsection{Point Process Models}
Point processes \cite{tppintro} are useful tools for continuous spatio-temporal event prediction, unaffected by granularity issues. By specifying continuous functions representing event occurrence rates, point process models can generate the distribution of future events. For this purpose, traditional point process models assume fixed functional forms with tunable parameters. Poison processes \cite{poisson}, Hawkes processes \cite{hawkes}, and self-correcting processes \cite{self-correct} formulate different forms of intensity functions conditioned on past events, accounting for the triggering effects between events. However, simple functional forms fail to capture complex event dependencies. Thus, neural networks have been widely applied to point process modeling problems. Most neural point process methods are merely sequential prediction models, including RNN-based methods \cite{rmtpp,nhp,lognormmix,unipoint}, transformer-based methods \cite{THP,SAHP,attnhp}, CNN-based methods \cite{ctpp,dltpp}, etc. Sequential models treat spatial features as discrete labels, lacking the ability to accurately predict spatial coordinates. Hence, STPP models have become a research hotspot in recent years. The core issue of STPP modeling is the normalization of the multi-dimensional spatio-temporal distribution. Most recent STPP methods adopt generative approaches to avoid the intractable normalization problem, including variational auto encoders (VAEs) \cite{deepstpp}, conditional normalising flows \cite{nstpp}, and diffusion-based models \cite{dstpp}.


\section{Preliminary}
\label{sec:preliminary}
\subsection{Spatio-temporal Point Processes}
\label{sec:STPP}
Spatio-temporal Point Processes (STPPs) are useful tools for modeling discrete event occurrences in continuous time and space. An STPP event sequence can be given as $ \mathcal{S}=\{(t_i,\boldsymbol{s}_i)\}_{i=1}^L $, where each event is characterized by a timestamp and a spatial coordinate. An STPP prediction problem can be formulated as fitting the joint distribution $ p(t,\boldsymbol{s}|\mathcal{H}_t) $, where $ \mathcal{H}_t=\{(t_j,\boldsymbol{s}_j)|t_j\leq t\} $ represents the history events that occurred before time $ t $. For simplicity, we use $ p^* $ to represent distributions that depend on the history $ \mathcal{H}_t $ hereafter.





\subsection{Neural Ordinary Differential Equations}
Neural Ordinary Differential Equations (ODEs) \cite{neuralode} have become a popular technique for modeling continuous dynamics. Using a neural network to specify the gradients of the dynamic variable at any point in its domain, we can establish a vector field to extrapolate the future evolution of the state by solving initial value problems.

\section{Methodology}
\label{sec:methodology}

\subsection{Overview}
As shown in Fig. \ref{fig:framework}, the model has an encoder-decoder architecture. The encoder simulates the global-local state evolution. Unlike previous works \cite{nstpp,njsde} that use a single global state vector, we propose to incorporate global and local dynamics by maintaining two types of dynamic states, including a location-independent global state $ \boldsymbol{z}^\mathbb{G} $ and $ K $ region-specific local states $ \boldsymbol{Z}^{\mathbb{L}}=[\boldsymbol{z}_1^\mathbb{L},\boldsymbol{z}_2^\mathbb{L},...,\boldsymbol{z}_K^\mathbb{L}] $ representing the dynamics of different spatial regions. 
We define two types of state evolution, namely extrapolations and jumps. The extrapolations model the smooth state transitions within event intervals, while the jumps simulate the abrupt state changes induced by event occurrences. Four networks are devised to model these two types of evolution of both global and local states. The encoder outputs are the trajectories of the global-local states along the temporal axis. Details about encoder networks can be found in subsection \ref{sec:evolution}. Note that local state evolution is based on a novel structure, named the Self-Adaptive Anchor Graph (SAAG), which captures complex spatial dependencies. We will introduce the mechanism of SAAG in subsection \ref{sec:SAAG}.

The decoder generates the predicted spatio-temporal distribution of the target event, i.e., $ p^*(\boldsymbol{s},t) $, by exploiting the information encoded in the global and local dynamic states. We decompose the joint distribution into two components as follows.
\begin{equation}
\label{eq:decomp}    
p^*(\boldsymbol{s},t)=p^*(t)\cdot p^*(\boldsymbol{s}|t)
\end{equation}
We model the temporal component $ p^*(t) $ with the conditional intensity function $ \lambda^*(t) $ using the following equation.
\begin{equation}
\label{eq:p-lambda}
    p^*(t)=\lambda^*(t)\cdot \exp(-\int_{{\Bar{t}}}^t \lambda^*(\tau)d\tau)
\end{equation}
The time-conditioned spatial distribution $ p^*(\boldsymbol{s}|t) $ is modelled with a Gaussian Mixture Model (GMM). Details about the decoder networks can be found in Subsection \ref{sec:dec}.

\vspace{-1em}
\subsection{Self-Adaptive Anchor Graph}
\label{sec:SAAG}
\begin{figure}[ht]
    \centering
    \vspace{-1em}
    \includegraphics[width=\linewidth]{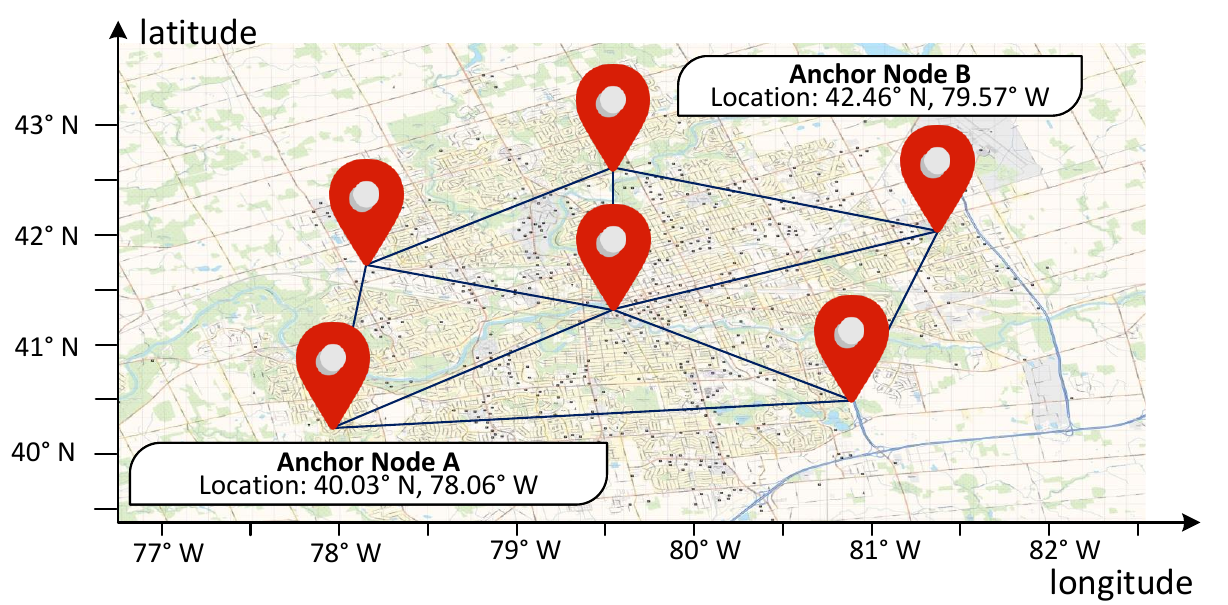}
    \vspace{-2.2em}
    \caption{An example of a spatially localised anchor graph. Each anchor node has a spatial coordinate. The nodes are connected with edges that represent the spatial correlations.}
    \label{fig:anchor}
    \vspace{-0.5em}
\end{figure}

An SAAG $ \mathcal{G} $ consists of a set of $ K $ anchor nodes and the correlation edges between them. Unlike the nodes in normal graphs, the anchor nodes are localized in the Euclidean space of event occurrences. Each of the anchor nodes is associated with a coordinate indicating its spatial location, that is, $ \boldsymbol{C}=[\boldsymbol{c}_1,...\boldsymbol{c}_K] $. The use of anchor nodes eliminates the need for explicit borders to split the space into regions. The anchor nodes are representatives of their nearby regions and store the event dynamics of the local areas. The locations of the anchor nodes are trainable in order to adaptively find the best localization points. Fig.\ref{fig:anchor} is an example of an anchor graph. 

Inspired by \cite{wavenet,agcrn,mtgnn,stemgnn,crossgnn,shen2024beyond}, we propose to use a self-adaptive adjacency learning approach to learn the hidden correlations between anchor nodes. We use a double-headed adjacency approach to represent the inter-region correlations, specifically a distance adjacency head and a latent adjacency head. The distance adjacency $ \boldsymbol{A}^d $ is built on the intuition that locations close in space should have a stronger correlation. We define the distance adjacency with the RBF kernel:
\vspace{-1em}

\begin{equation}
    \boldsymbol{A}^d[i,j]=
    \begin{cases}
        0, & i=j\\
        \exp(-\gamma\lVert\boldsymbol{c}_i-\boldsymbol{c}_j\rVert^2), & i \neq j
    \end{cases}
\end{equation}
where $ \gamma $ is a hyperparameter controlling the decay rate. However, there often exist hidden connections between locations that could not be expressed by Euclidean distance or any other prior knowledge we have. Thus, we also need to learn a latent adjacency to capture such hidden correlations. Inspired by \cite{mtgnn}, we propose to use a uni-directional adjacency learning approach defined as follows.
\begin{equation}
    \boldsymbol{A}^l=softplus\Bigl(\boldsymbol{E}_1\boldsymbol{E}_2^\mathrm{T}-\boldsymbol{E}_2\boldsymbol{E}_1^\mathrm{T}\Bigr)
\end{equation}
where $ \boldsymbol{E}_1,\boldsymbol{E}_2\in \mathbb{R}^{K\times d_E} $ are the trainable node embedding matrices.
This design guarantees that all relations are approximately uni-directional, i.e., if $ \boldsymbol{A}^l[i,j] $ has a relatively big value, $ \boldsymbol{A}^l[j,i] $ is guaranteed to be a small value close to zero. We use softplus activation instead of popular ReLU or LeakyReLU in order to maintain the smoothness of the neural ODE dynamics. We denote the double-headed adjacency matrix as $ \boldsymbol{A}=[\boldsymbol{A}^d;\boldsymbol{A}^l]\in \mathbb{R}^{2\times K\times K} $.

To fully exploit the spatial patterns learned by the graph, we devise two submodules used to encode the spatial information, namely the Location-aware GCN (L-GCN) and the Relative Location Encoder (RLE). 

\textbf{L-GCN} is our improved version of the vanilla GCN, which considers the relative positions of the anchor nodes during graph convolutions. Our purpose of devising L-GCN is to encode the inter-region depedencies learned by SAAG, where nodes are spatially localized. Unlike vanilla GCNs, where the edges are only associated with similarity weights, we also assign a vector to each edge indicating their relative positions.
\begin{equation}
    \boldsymbol{P}[i,j]=\tanh\Biggl(\boldsymbol{W}^\mathbb{P}_2\cdot\mathrm{SiLU}\bigl(\boldsymbol{W}^\mathbb{P}_1(\boldsymbol{c}_i-\boldsymbol{c}_j)+\boldsymbol{b}^\mathbb{P}_1\bigr)+\boldsymbol{b}^\mathbb{P}_2 \Biggr)
\end{equation}
where $ \boldsymbol{W}^\mathbb{P}_1 $, $ \boldsymbol{W}^\mathbb{P}_2 $, $ \boldsymbol{b}^\mathbb{P}_1 $ and $ \boldsymbol{b}^\mathbb{P}_2 $ are trainable parameters. Inspired by \cite{mtgnn}, we use a residual connection to keep a portion of the input information in each GCN layer to alleviate the over-smoothing problem. The computation of an L-GCN layer is formulated as follows.
\begin{equation}
\begin{split}
    \boldsymbol{H}^{(m)}[:, i]=&\beta \boldsymbol{H}^{(m-1)}[:, i]+\\
    &(1-\beta)\sum_{j=1}^N\Tilde{\boldsymbol{A}}[:, j,i]\Bigl(\boldsymbol{P}[j,i]\odot \boldsymbol{H}^{(m-1)}[:, j]\Bigr)
\end{split}
\end{equation}
where $ \boldsymbol{H}^{(m)}\in \mathbb{R}^{2\times K\times d} $ represents the double-headed hidden state at the $ m $-th L-GCN layer, $ \beta $ is a hyperparameter controlling the rate of state preservation, and $ \Tilde{\boldsymbol{A}}\in \mathbb{R}^{2\times K\times K} $ is the normalised adjacency matrix. Note that $ \boldsymbol{P} $ is used as a filter to reshape the message passed through the edges of the graph. The final output of L-GCN is aggregated by an information selection layer similar to \cite{mtgnn} as follows.
\vspace{-.5em}
\begin{equation}
    \boldsymbol{H}_{\mathrm{out}}=\sum_{j=1}^2\sum_{m=0}^M \boldsymbol{W}^{(m)}_j\boldsymbol{H}^{(m)}[j]+\boldsymbol{b}
\vspace{-.5em}
\end{equation}
which is a linear combination of the hidden states at all layers captured by both adjacency heads. L-GCN models the message-passing mechanism between different spatial regions, which encodes the spatial dependencies for local state evolution.

\textbf{RLE} is used to encode the effect an event occurrence cast on each of the anchor nodes, which can be used for local state update. For the $i$-th anchor node, given an event occurrence $ (t,\boldsymbol{s}) $, RLE encodes the relative direction $ \boldsymbol{\alpha}_i $ and the relative distance $ l_i $ of the event, with respect to the node position $ \boldsymbol{c}_i $. The event encoding with respect to the anchor node $ i $ is computed as follows.
\vspace{-1em}
\begin{equation}
    \boldsymbol{x}_i=\mathrm{MLP}(\boldsymbol{\alpha}_i)\cdot\exp\biggl(-\boldsymbol{\psi}\cdot l_i\biggr)
\end{equation}
where $ \boldsymbol{\alpha}_i=\frac{ \boldsymbol{s}-\boldsymbol{c}_i}{\lVert \boldsymbol{s}-\boldsymbol{c}_i\rVert_2} $ is a unit vector that represents the relative direction, and $ l_i=\lVert \boldsymbol{s}-\boldsymbol{c}_i\rVert_2 $ is the relative distance. The MLP transforms the direction vector into the hidden space. The relative distance $ l_i $ is used as the factor of an exponential decay, which is based on the intuition that the anchor nodes closer to the event should receive greater influence. $ \boldsymbol{\psi}\in \mathbb{R}^{d} $ is a trainable vector that controls the decay rate.

SAAG is the foundation of the proposed GSTPP model. Its ability to learn adaptive localization and correlation facilitates the capture of complex spatial dependencies. 

\subsection{Global-local State Evolution}
\label{sec:evolution}
The global-local state evolution is modelled with neural ODEs with jumps. The evolution of the global state $ \boldsymbol{z}^\mathbb{G}\in\mathbb{R}^{d_\mathrm{model}} $ and the local states $ \boldsymbol{Z}^\mathbb{L}\in\mathbb{R}^{K\times d_\mathrm{model}} $ are modeled simultaneously with two types of encoders. The extrapolation encoders model the smooth transition of the states within event intervals, while the jump encoders simulate the abrupt state changes induced by event occurrences.

\subsubsection{State Extrapolations}
The extrapolation encoders model the drift functions of ODEs. Specifically, given the global extrapolation encoder $ f^\mathbb{G} $ and the local extrapolation encoder $ f^\mathbb{L} $, the global-local state extrpolation can be formulated as follows.
\begin{equation}
\label{eq:global-ode}
    d\boldsymbol{z}^\mathbb{G}(t)=f^\mathbb{G}\biggl(\boldsymbol{z}^\mathbb{G}(t),t\biggr)dt
\end{equation}
\vspace{-1.5em}
\begin{equation}
\label{eq:local-ode}
    d\boldsymbol{Z}^\mathbb{L}(t)=f^\mathbb{L}\biggl(\boldsymbol{Z}^\mathbb{L}(t),t,\mathcal{G}\biggr)dt
\end{equation}
Given the initial state $ [\boldsymbol{z}^\mathbb{G}(t_0),\boldsymbol{Z}^\mathbb{L}(t_0)] $, we can compute the state at any time $ t $ before the next event occurrences by solving an initial value problem. Note that the evolution of the local states depends on the anchor graph $ \mathcal{G} $, because we associate each of the local states with an anchor node in the graph, and thus the local dynamics rely on the spatial correlations obtained by SAAG. 

The internal structures of the global and local extrapolation encoders are shown in Fig.\ref{fig:framework}. Inspired by \cite{gru-ode}, we adopt a GRU-based structure with residual subtraction at the end to ensure the numerical stability of the states. However, we use Time-dependent GRUs (T-GRUs) instead of vanilla ones for state updating to take into account the current time.  The spatial information learnt by SAAG is exploited by the local extrapolation encoder with the L-GCN network that encodes inter-region correlations.

\subsubsection{State Jumps}
The jump encoders simulate the abrupt state changes induced by event occurrences. The global-local dynamic states are updated with the information contained in the new event occurrences. The internal structures of the jump encoders are shown in Fig.\ref{fig:framework}. Specifically, the global jump encoder $ J^\mathbb{G} $ is simply a T-GRU network that takes the current location as input. The local jump encoder $ J^\mathbb{L} $ first transforms the event location into feature vectors corresponding to each anchor node using RLE, before passing them to T-GRU as input.




\subsection{Spaio-temporal Distribution Generation}
\label{sec:dec}
The joint distribution of the target event $ p^*(\boldsymbol{s},t) $ can be decomposed into temporal and spatial components as given in Eq.\ref{eq:decomp}. Thus, we propose to use two decoders to generate the two components, respectively.

The temporal decoder predicts the conditional intensity function $ \lambda^*(t) $ using the global state $ \boldsymbol{z}^\mathbb{G} $. We formulate the temporal decoding network as follows.
\vspace{-0.5em}
\begin{equation}
    \lambda^*(t)=softplus\biggl(\mathrm{MLP}\bigl(\boldsymbol{z}^\mathbb{G}(t)\bigr)\biggr)
\end{equation}
where the softplus activation guarantees the non-negativity of the function. The conditional temporal pdf $ p^*(t) $ can be approximated with Eq. \ref{eq:p-lambda}.

The spatial decoder predicts the conditional spatial distribution of the event at the given time $ t $, i.e., $ p^*(\boldsymbol{s}|t) $, from the local states $ \boldsymbol{Z}^\mathbb{L}(t) $. The spatial distribution is formulated as a mixture distribution as follows.
\begin{equation}
    p^*(\boldsymbol{s}|t)=\sum_{i=1}^K\gamma_ip_i^*(\boldsymbol{s}|t)
\end{equation}
where $ \gamma_i $ is the mixture coefficient obtained as:
\begin{equation}
    \gamma_i=softmax\biggl(\mathrm{MLP}(\boldsymbol{z}_i^\mathbb{L}(t))\biggr)
\end{equation}
The $ i $-th mixture component $ p_i^*(\boldsymbol{s}|t) $ is a Gaussian distribution generated from the $ i $-th anchor node in SAAG.
\vspace{-1em}
\begin{equation}
    p_i^*(\boldsymbol{s}|t)=\mathcal{N}\biggl(\boldsymbol{s};\boldsymbol{\mu}\bigl(\boldsymbol{z}_i^\mathbb{L}(t),\boldsymbol{c}_i\bigr),\boldsymbol{\sigma}^2\bigl(\boldsymbol{z}_i^\mathbb{L}(t)\bigr)\biggr)
\end{equation}
where the mean network takes into account the current local state and location of the corresponding anchor node.
\vspace{-1em}
\begin{equation}
    \boldsymbol{\mu}\biggl(\boldsymbol{z}_i^\mathbb{L}(t,\boldsymbol{c}_i)\biggr)=\mathrm{MLP}\biggl(\boldsymbol{z}_i^\mathbb{L}(t)\biggr)+\boldsymbol{c}_i
\end{equation}
Note that the anchor coordinate is added to the MLP output to retain the anchors' locality. The variance network $ \boldsymbol{\sigma}^2(\boldsymbol{z}_i^\mathbb{L}(t)) $ is formulated as:
\begin{equation}
    \boldsymbol{\sigma}^2\biggl(\boldsymbol{z}_i^\mathbb{L}(t)\biggr)=\exp\biggl(\mathrm{MLP}\bigl(\boldsymbol{z}_i^\mathbb{L}(t)\bigr)\biggr)
\end{equation}
The exponential activation guarantees the non-negativity of the output. By combining the result of temporal and spatial decoders according to Eq. \ref{eq:decomp}, we obtain the joint distribution $ p^*(\boldsymbol{s},t) $.

\subsection{Maximum Log-likelihood Estimation}
We propose to train our GSTPP with a Maximum Log-likelihood Estimation (MLE) approach. The training objective is the log-likelihood of the spatio-temporal event sequence $ \mathcal{S}={(t_i,\boldsymbol{s}_i)}_{i=1}^N $.
\begin{equation}
    \log p_{\theta}(\mathcal{S})=\sum_{i=1}^N \log p^*_\theta(t_i,\boldsymbol{s}_i)
\end{equation}
where $ \theta $ represents all the trainable parameters of the encoder-decoder network. The training process hence generalizes to the following optimization problem.
\begin{equation}
    \max_{\theta} \log p_\theta(\mathcal{S})
\end{equation}
In practice, we use the back propagation and gradient descent methods to learn the model parameters. The adjoint sensitivity method is also adopted to address the efficiency issue of ODEs.

\section{Experiments}
\label{sec:experiments}

\subsection{Datasets}
\begin{figure}[H]
\centering
\vspace{-2em}
\subfloat[Number of event sequences]{\includegraphics[width=1.7in]{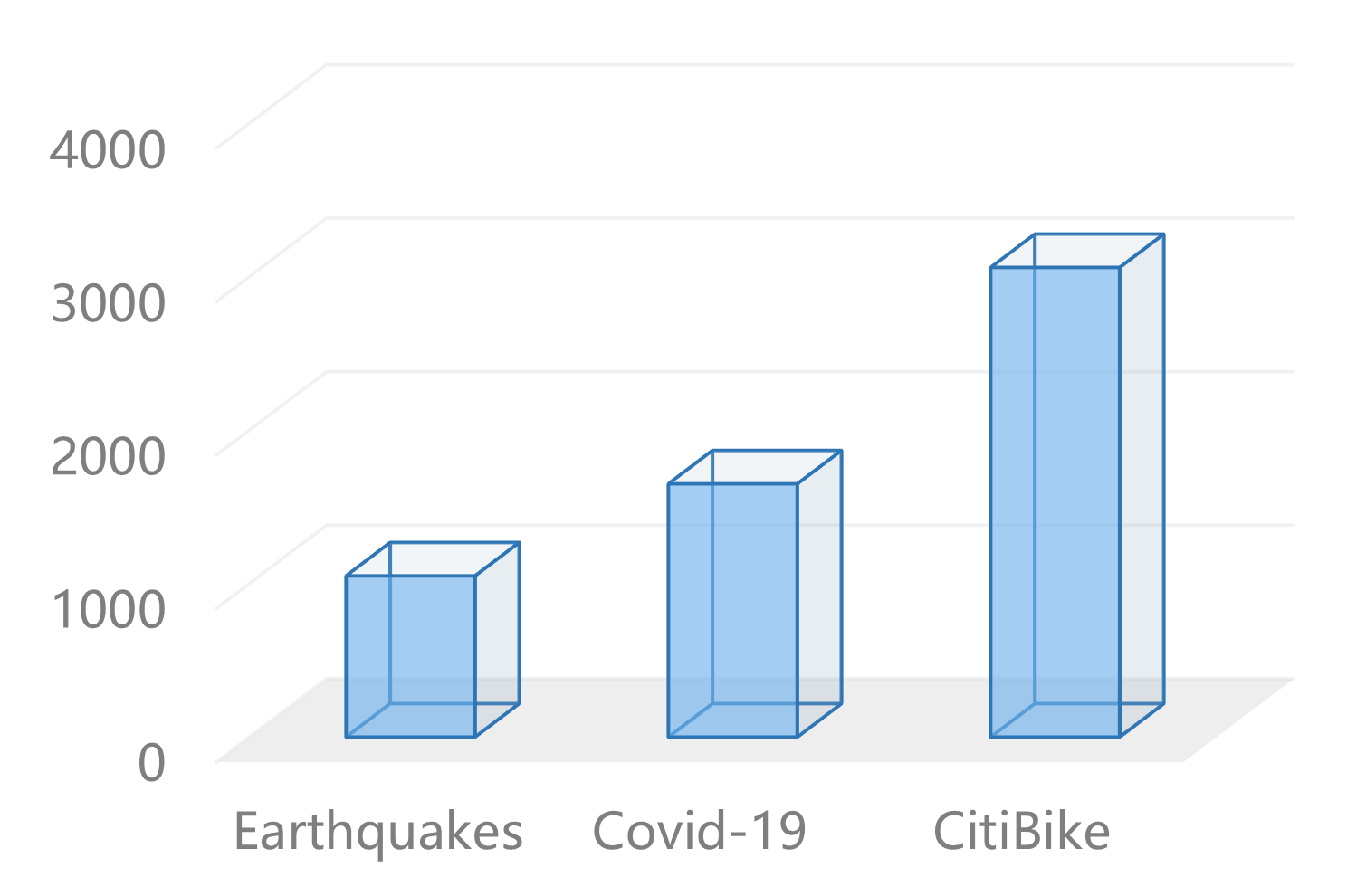}}
\subfloat[Average sequence length]{\includegraphics[width=1.7in]{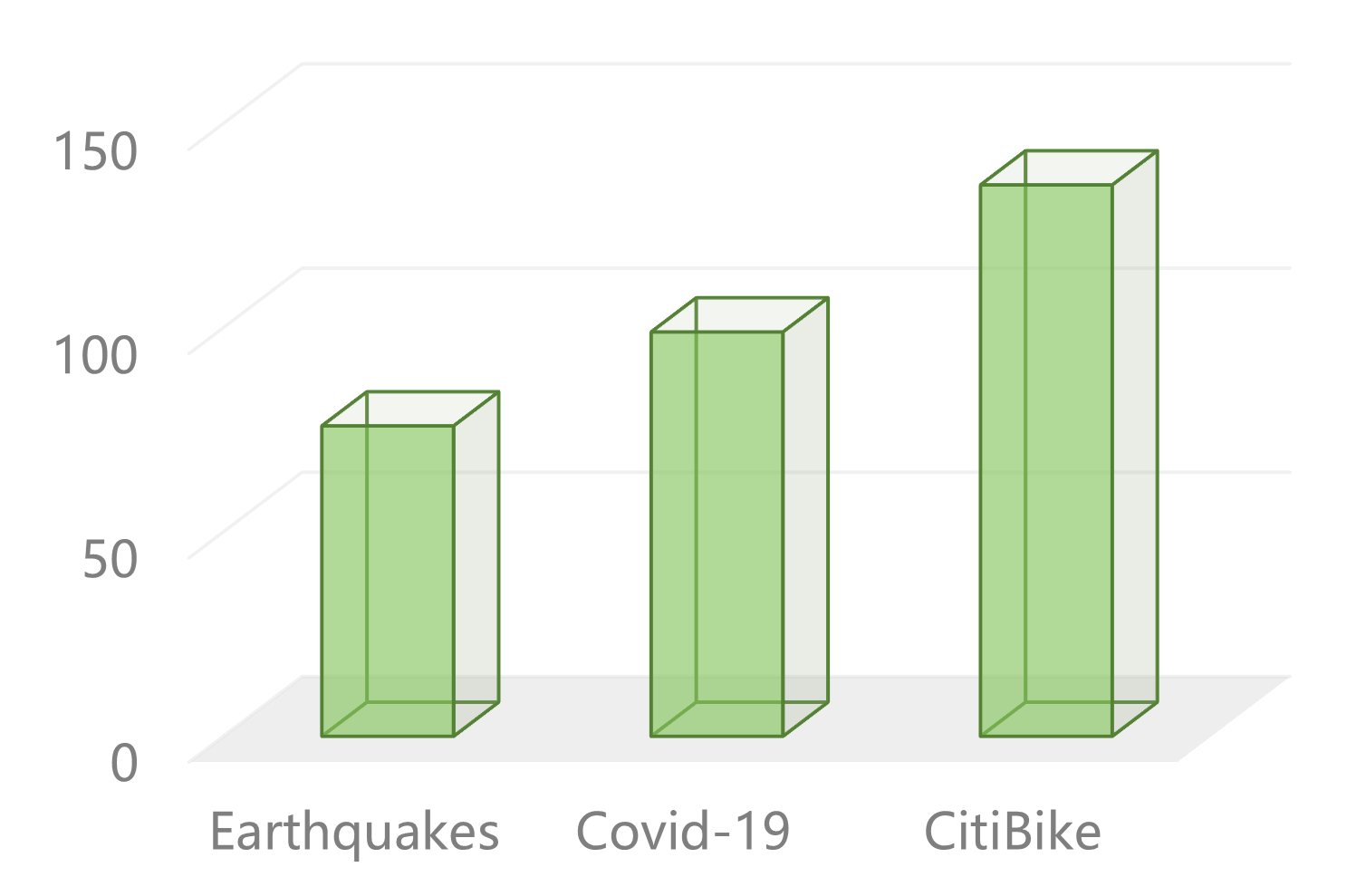}}
\caption{The statistics of the three datasets.}
\vspace{-2em}
\label{fig:dataset}
\end{figure}

We adopt three real-world spatio-temporal datasets for model evaluation and comparison, namely \textit{Earthquakes} \cite{earthquake-dataset}, \textit{COVID-19} \cite{covid-dataset}, and \textit{CitiBike}. Fig.\ref{fig:dataset} shows the total sequence numbers and average sequence lengths of the three datasets. Using datasets of different scales helps us to understand our model's performance in different situations. 
Please refer to Appendix \ref{appx:dataset} for an introduction of the three datasets.




\begin{table*}[!ht]
  \caption{Probabilistic evaluation results} \label{tab:nll}
  \vspace{-1em}
  \resizebox{2\columnwidth}{!}{
  \begin{tabular}{lrrrrrrrrr}
    \toprule
    & \multicolumn{3}{c}{\texttt{Earthquakes}}   & \multicolumn{3}{c}{\texttt{COVID-19}} & \multicolumn{3}{c}{\texttt{CitiBike}}\\
    \cmidrule(lr){2-4} \cmidrule(lr){5-7} \cmidrule(lr){8-10}
    Methods & \multicolumn{1}{c}{ST-NLL }      & \multicolumn{1}{c}{T-NLL}   & \multicolumn{1}{c}{S-NLL}          & \multicolumn{1}{c}{ST-NLL}      & \multicolumn{1}{c}{T-NLL}   & \multicolumn{1}{c}{S-NLL}
    & \multicolumn{1}{c}{ST-NLL}      & \multicolumn{1}{c}{T-NLL}   & \multicolumn{1}{c}{S-NLL}\\
\midrule
RMTPP& \multicolumn{1}{c}{-} & -0.210{\tiny$\pm$0.008} & \multicolumn{1}{c}{-} &  \multicolumn{1}{c}{-} & -2.426{\tiny$\pm$0.003} & \multicolumn{1}{c}{-} & \multicolumn{1}{c}{-} & -1.107{\tiny$\pm$0.001} & \multicolumn{1}{c}{-}\\
NHP & \multicolumn{1}{c}{-} & -0.198{\tiny$\pm$0.001} & \multicolumn{1}{c}{-} &  \multicolumn{1}{c}{-} & -2.229{\tiny$\pm$0.013} & \multicolumn{1}{c}{-} & \multicolumn{1}{c}{-} & -1.030{\tiny$\pm$0.015} & \multicolumn{1}{c}{-}\\
LogNormMix & \multicolumn{1}{c}{-} & -0.260{\tiny$\pm$0.000} & \multicolumn{1}{c}{-} &  \multicolumn{1}{c}{-} & -2.430{\tiny$\pm$0.000} & \multicolumn{1}{c}{-} & \multicolumn{1}{c}{-} & -1.114{\tiny$\pm$0.000} & \multicolumn{1}{c}{-}\\
CTPP & \multicolumn{1}{c}{-} & -0.234{\tiny$\pm$0.005} & \multicolumn{1}{c}{-} &  \multicolumn{1}{c}{-} & -2.431{\tiny$\pm$0.003} & \multicolumn{1}{c}{-} & \multicolumn{1}{c}{-} & -1.118{\tiny$\pm$0.004} & \multicolumn{1}{c}{-}\\
\midrule
Conditional-KDE & \multicolumn{1}{c}{-} & \multicolumn{1}{c}{-} & 5.859{\tiny$\pm$0.001} &  \multicolumn{1}{c}{-} &\multicolumn{1}{c}{-} & 0.646{\tiny$\pm$0.000} & \multicolumn{1}{c}{-} & \multicolumn{1}{c}{-} & -4.540{\tiny$\pm$0.000}\\  
\midrule
NJSDE & 5.066{\tiny$\pm$0.013} & -0.186{\tiny$\pm$0.005} & 5.252{\tiny$\pm$0.012}  & -1.974{\tiny$\pm$0.006} & -2.251{\tiny$\pm$0.004} & 0.277{\tiny$\pm$0.005} &  -5.757{\tiny$\pm$0.002} & -1.092{\tiny$\pm$0.002} & -4.665{\tiny$\pm$0.001}\\ 
NSTPP-Jump & 4.568{\tiny$\pm$0.004} & -0.245{\tiny$\pm$0.002} & 4.813{\tiny$\pm$0.004}  & -2.414{\tiny$\pm$0.005} & -2.431{\tiny$\pm$0.001} & 0.017{\tiny$\pm$0.005} & -6.138{\tiny$\pm$0.002}& -1.113{\tiny$\pm$0.001}& -4.981{\tiny$\pm$0.001}\\
NSTPP-Attn & \underline{4.481{\tiny$\pm$0.008}} & \underline{-0.264{\tiny$\pm$0.003}} & \underline{4.745{\tiny$\pm$0.005}}  & \underline{-2.445{\tiny$\pm$0.005}}& \underline{-2.432{\tiny$\pm$0.002}} & \underline{-0.013{\tiny$\pm$0.003}} & \underline{-6.289{\tiny$\pm$0.002}} & \underline{-1.118{\tiny$\pm$0.000}} & \underline{-5.171{\tiny$\pm$0.002}}\\
DSTPP & 5.946{\tiny$\pm$0.036} & 0.306{\tiny$\pm$0.004} & 5.640{\tiny$\pm$0.040}  & -1.385{\tiny$\pm$0.008} & -2.084{\tiny$\pm$0.004} & 0.699{\tiny$\pm$0.004} & -5.138{\tiny$\pm$0.004} & -0.785{\tiny$\pm$0.002} & -4.353{\tiny$\pm$0.002}\\

\midrule
\textbf{GSTPP} & \textbf{3.503{\tiny$\pm$0.005}}& \textbf{-0.265{\tiny$\pm$0.001}}& \textbf{3.768{\tiny$\pm$0.004}}&  \textbf{-2.465{\tiny$\pm$0.007}}& \textbf{-2.432{\tiny$\pm$0.001}}& \textbf{-0.034{\tiny$\pm$0.008}}& \textbf{-6.440{\tiny$\pm$0.003}}& \textbf{-1.119{\tiny$\pm$0.000}} & \textbf{-5.321{\tiny$\pm$0.003}}\\
\bottomrule
\end{tabular}
    }
\end{table*}

\begin{table*}[!ht]
\vspace{-1em}
\centering
  \caption{Sampling evaluation results} \label{tab:sample}
  \vspace{-1em}
  \resizebox{1.4\columnwidth}{!}{
  \begin{tabular}{lrrrrrr}
    \toprule
    & \multicolumn{2}{c}{\texttt{Earthquakes}}   & \multicolumn{2}{c}{\texttt{COVID-19}} & \multicolumn{2}{c}{\texttt{CitiBike}}\\
    \cmidrule(lr){2-3} \cmidrule(lr){4-5} \cmidrule(lr){6-7}
    Methods & \multicolumn{1}{c}{T-RMSE }      & \multicolumn{1}{c}{S-Dist}             & \multicolumn{1}{c}{T-RMSE}      & \multicolumn{1}{c}{S-Dist}   
    & \multicolumn{1}{c}{T-RMSE}      & \multicolumn{1}{c}{S-Dist}   \\
\midrule

LogNormMix & 0.703{\tiny$\pm$0.005} & \multicolumn{1}{c}{-} & 0.132{\tiny$\pm$0.002} & \multicolumn{1}{c}{-} & 0.581{\tiny$\pm$0.008} & \multicolumn{1}{c}{-}\\
CTPP & 0.668{\tiny$\pm$0.005} & \multicolumn{1}{c}{-} & 0.124{\tiny$\pm$0.002} & \multicolumn{1}{c}{-} & 0.590{\tiny$\pm$0.006 } & \multicolumn{1}{c}{-}\\
\midrule
Conditional KDE & \multicolumn{1}{c}{-} & 11.315{\tiny$\pm$0.658} & \multicolumn{1}{c}{-} & 0.688{\tiny$\pm$0.047} & \multicolumn{1}{c}{-} & 0.718{\tiny$\pm$0.001}\\
\midrule
NSTPP-Attentive & \textbf{0.647{\tiny$\pm$0.010}}& \underline{8.971{\tiny$\pm$0.020}}& \underline{0.110{\tiny$\pm$0.004}}& 0.581{\tiny$\pm$0.010}& \underline{0.301{\tiny$\pm$0.005}}& 0.045{\tiny$\pm$0.010}\\
DSTPP & 0.659{\tiny$\pm$0.005} & 8.668{\tiny$\pm$0.124} & \textbf{0.109{\tiny$\pm$0.001}}& \underline{0.567{\tiny$\pm$0.013}}& \textbf{0.296{\tiny$\pm$0.005}}& \underline{0.045{\tiny$\pm$0.002}}\\

\midrule
\textbf{GSTPP} & \underline{0.652{\tiny$\pm$0.002}}& \textbf{6.970{\tiny$\pm$0.150}}& 0.116{\tiny$\pm$0.002} &  \textbf{0.416{\tiny$\pm$0.005}}& 0.497{\tiny$\pm$0.021}& \textbf{0.034{\tiny$\pm$0.010}}\\
\bottomrule
\end{tabular}
    }
\vspace{-1.5em}
\end{table*}

\subsection{Training Details}
The proposed GSTPP is implemented using Python with the Google JAX framework. The implementation code is available online\footnote{\url{https://github.com/AnthonyChouGit/GSTPP}}. Diffrax \cite{diffrax} is used to accelerate the approximation and back-propagation of neural ODEs. All models used for comparison are trained with the AdamW optimizer provided. The learning rate is initially set to 1e-3 and reduced using a sinusoidal decaying strategy through training. All experiments are performed on a workstation equipped with an Intel Xeon Gold 5218 CPU and NVIDIA GeForce RTX 4090 GPU.

\vspace{-0.5em}
\subsection{Baselines}
We used nine baselines for the experimental comparison to validate the superiority of the proposed GSTPP, including four purely temporal models (RMTPP \cite{rmtpp}, NHP \cite{nhp}, LogNormMix \cite{lognormmix}, and CTPP \cite{ctpp}), one purely spatial model (conditional KDE \cite{kde}), and four spatio-temporal models (NJSDE \cite{njsde}, NSTPP-Attn, NSTPP-Jump \cite{nstpp} and DSTPP \cite{dstpp}).
Please refer to Appendix \ref{appx:baseline} for a brief introduction of the baselines.

\vspace{-0.5em}
\subsection{Probabilistic Evaluation}

The Negative Log-Likelihood (NLL) is one of the most widely used metrics to evaluate probabilistic modeling, which measures how well the predicted distribution fits the real samples. For STPP modeling, the value of spatio-temporal NLL (ST-NLL) can be decomposed into the sum of temporal NLL (T-NLL) and spatial NLL (S-NLL). Table \ref{tab:nll} compares the NLL results between the proposed GSTPP and the baseline models. Lower values in the table indicate better performance. Note that the NLL results of DSTPP are approximated using the Evidence Lower BOund (ELBO) of the DDPM. Pure temporal models do not have S-NLL results because they cannot make fine-grained spatial predictions. Similarly, the purely spatial model Conditional-KDE cannot predict temporal distributions, thus its T-NLL results are not presented. The best results for each metric are bolded, and the second-best ones are underlined. Several observations can be made from the results.

First, with proper adjustments, joint modeling of temporal and spatial event distributions can improve overall performance. Although temporal and spatial distributions can be modeled separately using one purely temporal model like LogNormMix and one purely spatial model like Conditional-KDE, the overall prediction performance does not match joint STPP models like NSTPP and GSTPP. This is because spatial and temporal event features are mutually dependent, thus joint spatio-temporal modeling better encodes their internal correlations, leading to better overall performance.


Second, continuous dynamic state simulation is crucial to accurate probabilistic event prediction. As can be seen in the table, DSTPP exhibits uncompetitive NLL results compared to other spatio-temporal models. This is because NJSDE, NSTPP and GSTPP all model the continuous state transitions with differential equations, which enables the system to perceive the infinitesimal distributional change induced by the passing of time and event occurrences. However, DSTPP takes the event times and locations as a simple time series, lacking the ability to understand the complex and fine-grained dynamics of the target distribution.

Third, the proposed GSTPP is consistently superior to other STPP models in both temporal and spatial probabilistic prediction. As can be seen in the table, the proposed GSTPP surpasses all baseline models on all metrics. NSTPP-Attn is the strongest baseline and is second in all metrics. GSTPP's advantage in T-NLL is slight because it adopts a similar temporal intensity approximation approach as NSTPP with minor modifications. However, its superiority in spatial modeling is quite significant. This can be explained by GSTPP's ability to encode the spatial correlations and heterogeneity of different spatial regions, attributed to our novel self-adaptive anchor graph.

\vspace{-0.5em}
\subsection{Sampling Evaluation}
Aside from probabilistic evaluation, the sample quality is also a very important performance indicator of STPP models. We measure the sample quality with two evaluation metrics, namely T-RMSE and S-Dist. T-RMSE stands for the root mean squared error of the time samples of the predicted events. S-Dist stands for the average Euclidean distance between the sampled and real locations. We compare GSTPP's sample quality against 5 baselines (see Tab. \ref{tab:sample}). A few previously mentioned baselines are excluded here because they either prohibit efficient sampling or exhibit incompetitive results.

Similar to probabilistic results, joint spatio-temporal models generally presents better sampling results than purely temporal or spatial models due to the internal dependency between time and space. DSTPP presents strong sampling performance in spite of its failure in probabilistic prediction. The proposed GSTPP cannot beat the two spatio-temporal baselines in terms of T-RMSE, but the performance gap is very small and tolerable. However, GSTPP's spatial sampling performance surpasses all baselines with a significant margin, which again justifies the contribution of our novel self-adaptive anchor graph.

\vspace{-1em}
\subsection{Parameter Analysis}
\begin{figure}[H]
\vspace{-2em}
\centering
\subfloat[Earthquakes]{\includegraphics[width=0.33\columnwidth]{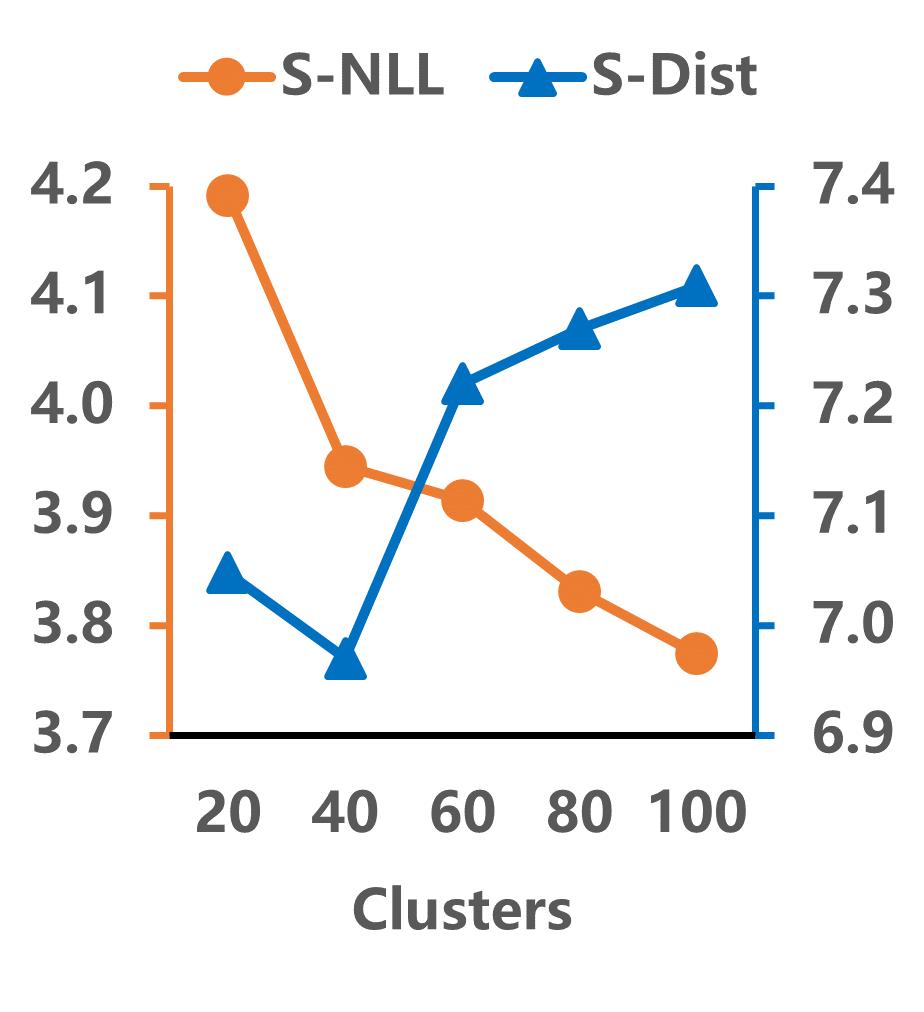}}
\subfloat[COVID-19]{\includegraphics[width=0.33\columnwidth]{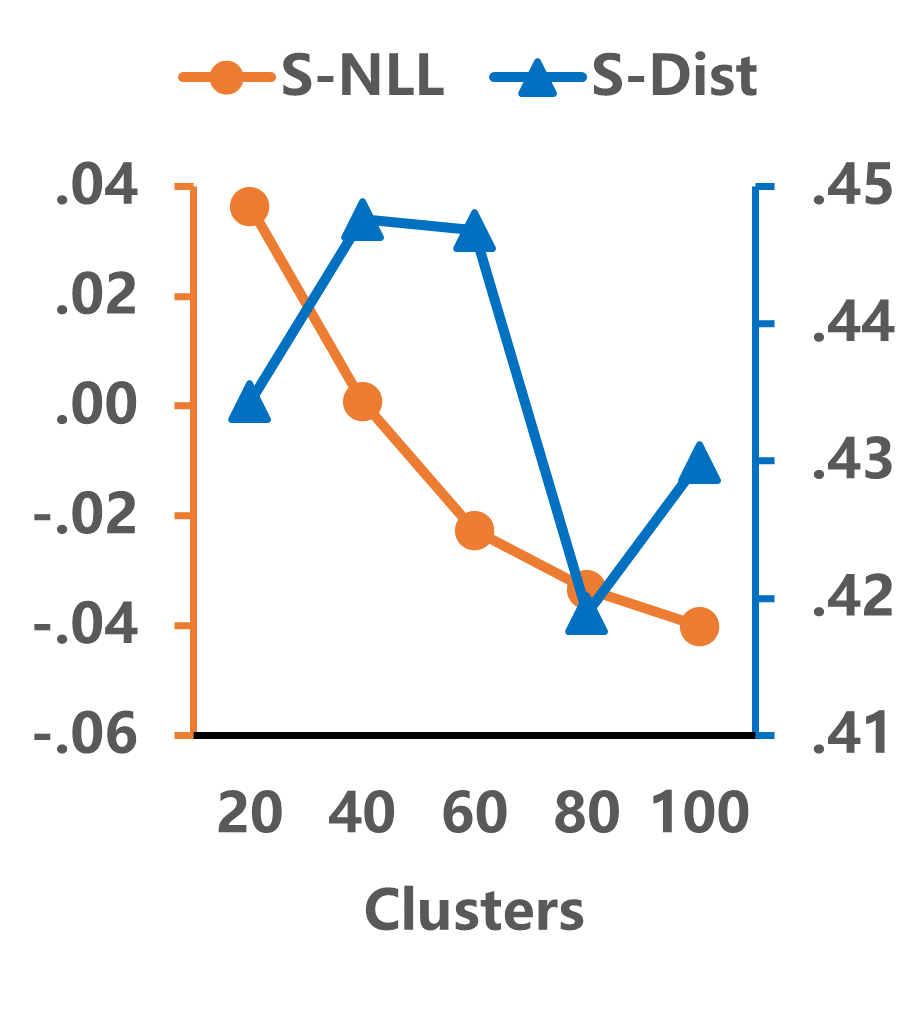}}
\subfloat[CitiBike]{\includegraphics[width=0.34\columnwidth]{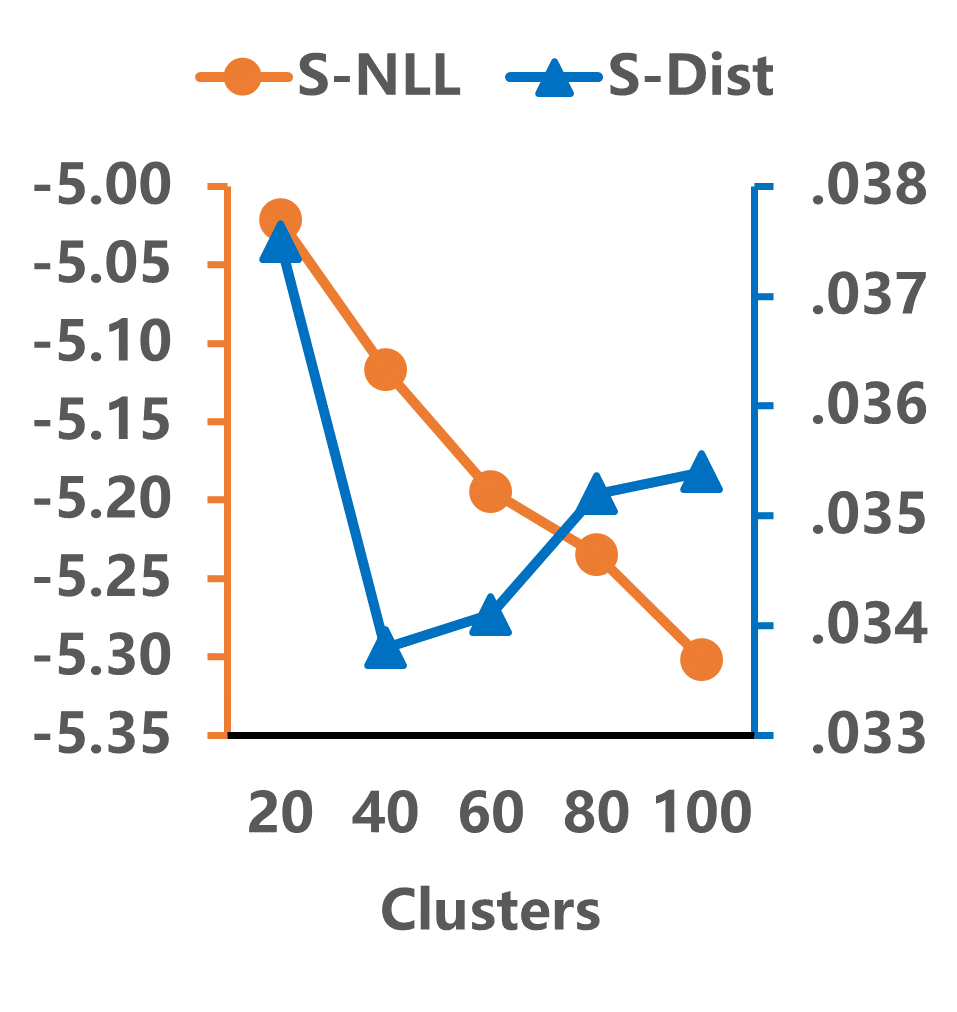}}
\vspace{-0.5em}
\caption{The spatial probabilistic and sampling performance of GSTPP using different number of anchor nodes (clusters).}
\vspace{-1em}
\label{fig:clusters}
\end{figure}

The number of anchor nodes is essential to the performance of GSTPP. To find out the sensitivity of GSTPP to this hyperparameter, we tested S-NLL and S-Dist of GSTPP using a different number of clusters, specifically 20, 40, 60, 80 and 100. The results are shown in Fig. \ref{fig:clusters}. 

We observe that the S-NLL values monotonically decrease as the number of clusters increases. However, for S-Dist, more clusters do not necessarily lead to better results. The models with a certain number of clusters (40 or 80 in our experiments) produce the best spatial sample quality. As the cluster number increases, the sample quality starts to deteriorate quickly.

\begin{figure}[H]
\vspace{-1.5em}
\centering
\subfloat[40 clusters]{\includegraphics[width=0.4\columnwidth]{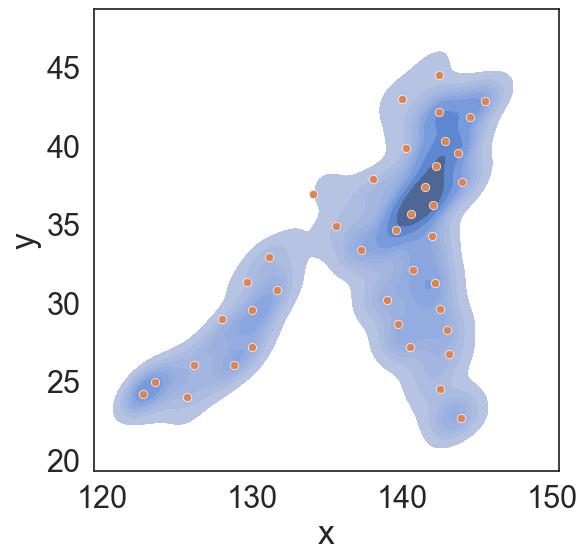}}
\hspace{.7em}
\subfloat[100 clusters]{\includegraphics[width=0.4\columnwidth]{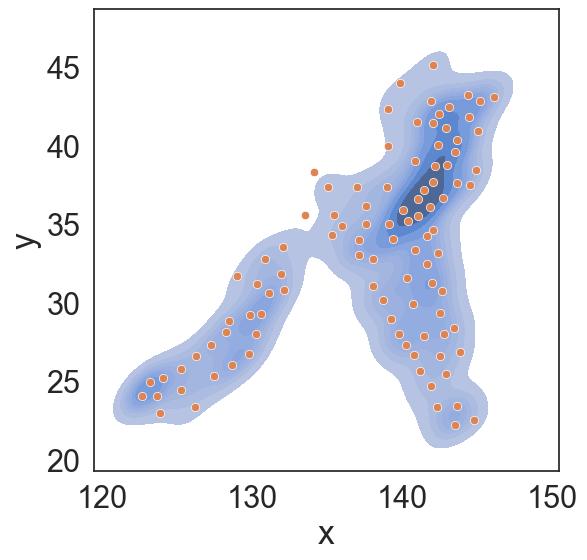}}
\vspace{-1em}
\caption{Overall spatial distribution and anchor positions trained using different numbers of clusters on \textit{Earthquakes} dataset. The x and y axis represent the longitude and latitude, respectively.}
\vspace{-1.5em}
\label{fig:cluster_vis}
\end{figure}

In Fig. \ref{fig:cluster_vis}, we visualize the distribution of the event locations together with the anchors trained with different numbers of clusters. As can be seen, anchors tend to locate themselves in areas where events are densely distributed. As the number of clusters increases, the anchors further populate dense areas and become closer to each other, while some even fall into the sparse areas.

More anchor nodes can obviously capture more complex spatial dependencies and produce more flexible and complex mixture distributions to approximate the real spatial distribution of events, which leads to better NLL results. However, with a large number of clusters, the anchors become too crowded, causing heavy overlaps between their mixture components. This makes the model generate over-complex distributions, which are more sensitive to noise and prone to overfitting.




\vspace{-.5em}
\subsection{Ablation Study}
\begin{figure}[H]
 \vspace{-1.5em}
\centering
\subfloat[Earthquakes]{\includegraphics[width=0.333\columnwidth]{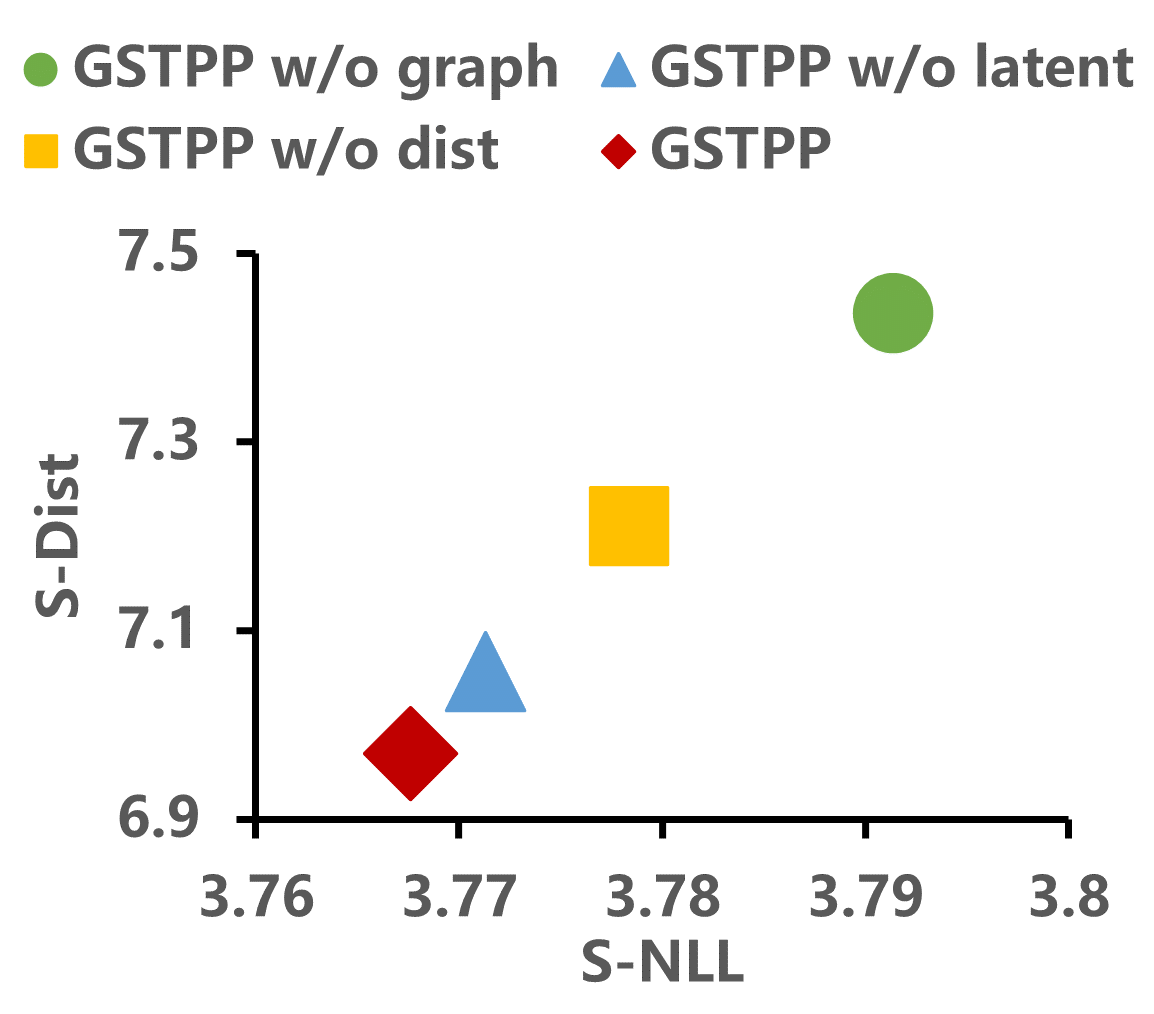}}
\subfloat[COVID-19]{\includegraphics[width=0.333\columnwidth]{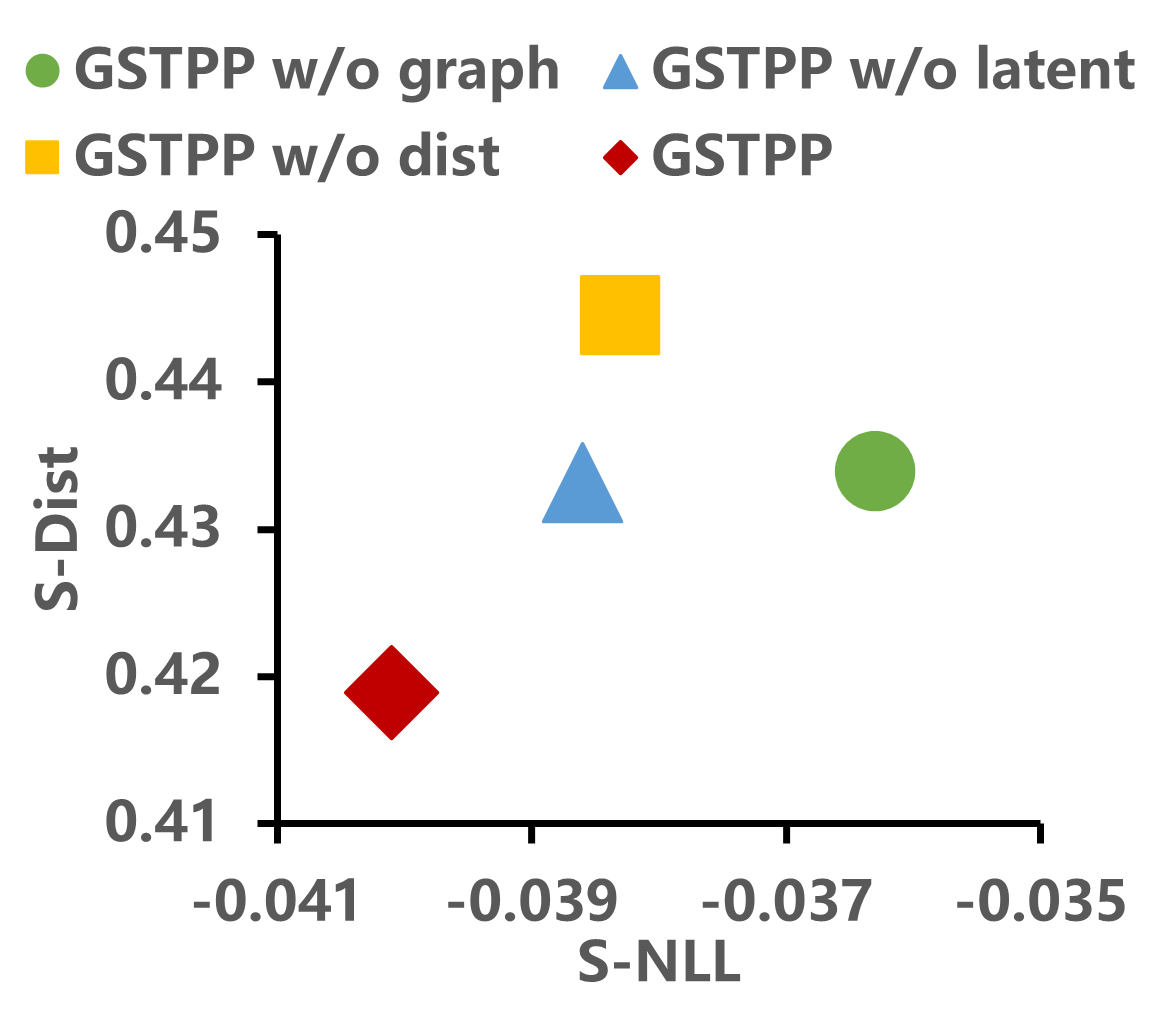}}
\subfloat[CitiBike]{\includegraphics[width=0.333\columnwidth]{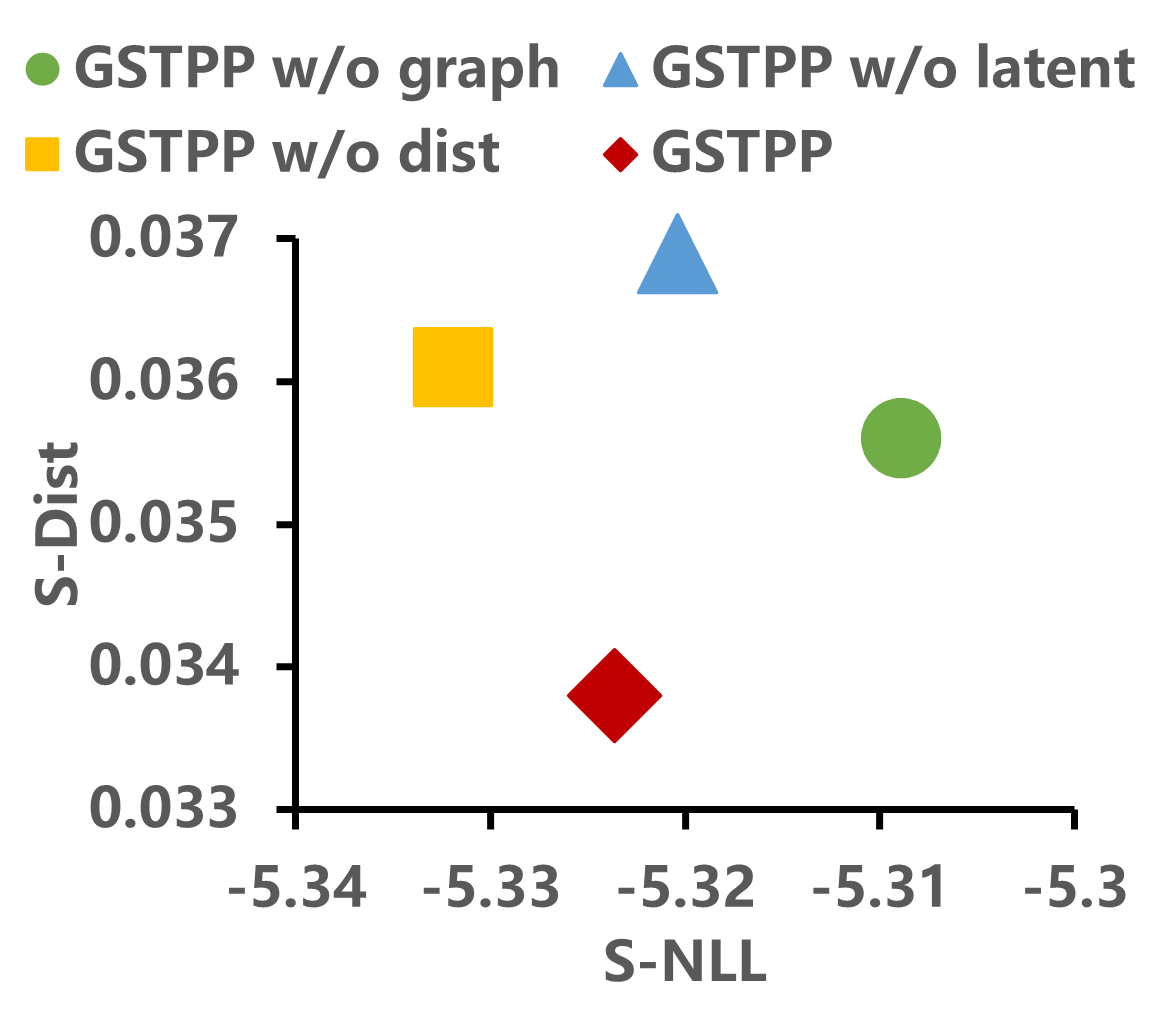}}
\vspace{-1em}
\caption{Spatial probabilistic and sampling performance of different model variants.}
\label{fig:ablation}
\vspace{-1.5em}
\end{figure}

The adaptive anchor graph used by GSTPP can capture the correlation and heterogeneity between different spatial regions and thus is crucial in improving the accuracy of the prediction of the location of the event. As mentioned in section \ref{sec:methodology}, we use two different graphs in the model to construct the spatial message-passing mechanism, namely the latent graph and the distance graph. The former is trainable and captures latent correlations between locations, while the latter focuses on distance-related relationships. We implemented three simplified versions of GSTPP to validate the superiority of our design. ``GSTPP w/o dist" stands for the model variant that does not have a distance graph. Similarly, ``GSTPP w/o latent" is deprived of the latent graph and only depends on the pairwise distance of the anchors for message passing. ``GSTPP w/o graph" has no graphs at all and the hidden states of all anchor nodes evolve independently without message passing. Fig. \ref{fig:ablation} shows the comparison of these model variants based on S-NLL and S-Dist. 

As can be seen in the figure, GSTPP generally outperforms all its simplified variants on the three datasets, which means that the proposed GSTPP structure is reasonable and superior. ``GSTPP w/o graph" has the generally worst performance on all three datasets, indicating that ignoring correlations between different spatial regions greatly compromises prediction accuracy. Another observation is that different datasets are sensitive to different graphs. ``GSTPP w/o latent" beats ``GSTPP w/o dist" on \textit{Earthquakes} and \textit{COVID-19}, which means the distance graph contributes more to the spatial correlation in these two settings. In contrast, the latent graph dominates the spatial correlation in \textit{CitiBike}, indicating that the pattern of event occurrences in this setting is less distance-related.

\section{Conclusion}
In this work, we propose a novel GSTPP framework that promotes the performance of fine-grained spatio-temporal event prediction. We address the issue of spatial heterogeneity and correlations between different regions that greatly affect event occurrences, which has never been considered by state-of-the-art methods. The proposed framework incorporates global and local state dynamics with a novel encoder-decoder architecture. We introduce a novel self-adaptive anchor graph to capture the complex spatial dependencies within the continuous spatial area. By leveraging the spatial patterns learned by the graph, the model can make more accurate predictions of future events. Extensive experiments demonstrate the effectiveness of the proposed framework and validate its advantages over state-of-the-art approaches.

\vspace{-.5em}

\section*{Acknowledgements}
This work was supported by the Natural Science Foundation of China under Grants 62376055 and 62276053. It was also partly supported by the Fundamental Research Funds for the Central Universities (No.ZYGX2021J019).


\bibliographystyle{siam}
\bibliography{bibitems}

\newpage
\section*{Appendix}
\appendix
\section{Datasets}
\label{appx:dataset}
\begin{itemize}
    \item \textbf{\textit{Earthquakes}} \cite{earthquake-dataset} dataset contains spatio-temporal records of earthquakes in Japan between 1990 and 2020 whose magnitude is no less than 2.5. The total number of event sequences is 1050. The average sequence length is 76.

    \item \textbf{\textit{COVID-19}}  dataset contains spatio-temporal records of COVID-19 cases in New Jersey released by the New York Times. We adopt the same pre-processing procedure as \cite{nstpp} to obtain fine-grained spatial coordinates. The total number of event sequences is 1650. The average sequence length is 99.

    \item \textbf{\textit{CitiBike}} dataset contains spatio-temporal records of trip starts of a bike-sharing service in New York City. The total number of event sequences is 3060. The average sequence length is 135.
\end{itemize}
\section{Baselines}
\label{appx:baseline}
\subsection{Temporal Baselines}
Purely temporal models focus on forecasting the trend of event occurrence rates. They can only take the spatial coordinates of events as normal feature vectors, completely ignoring spatial correlations, and cannot predict future event locations.

\begin{itemize}
    \item \textbf{RMTPP} \cite{rmtpp} is one of the earliest neural TPP models. They propose using an RNN to encode history and predict future intensity with exponential functions.

    \item \textbf{NHP} \cite{nhp} improves the flexibility of the modeling by devising a novel continuous-time LSTM architecture to better capture history event patterns.

    \item \textbf{LogNormMix} \cite{lognormmix} is an intensity-free TPP model that predicts future event probabilities using mixture distributions, greatly improving training efficiency and prediction accuracy.

    \item \textbf{CTPP} \cite{ctpp} incorporates local convolutional networks with global recurrent networks to fully encode short- and long-term event contexts.
\end{itemize}

\end{document}